\documentclass[conference]{IEEEtran}
\usepackage[latin9]{inputenc}
\usepackage{amsmath}
\usepackage{color}
\usepackage{graphicx}
\usepackage{multirow}
\usepackage{subfig}
\usepackage[dvipsnames]{xcolor}
\usepackage{makecell}

\ifCLASSINFOpdf

\else

\fi

\usepackage{stfloats}

\begin{document}

\title{CNN Detection of GAN-Generated Face Images based on Cross-Band Co-occurrences Analysis}

\author{\IEEEauthorblockN{*Mauro Barni }
\IEEEauthorblockA{University of Siena, Italy \\
Email: barni@diism.unisi.it}
\and
\IEEEauthorblockN{Kassem Kallas}
\IEEEauthorblockA{University of Siena, Italy\\
Email:  k\_kallas@hotmail.com}
\and
\IEEEauthorblockN{Ehsan Nowroozi}
\IEEEauthorblockA{University of Siena, Italy\\
Email:  ehsan.nowroozi65@gmail.com}
\and
\IEEEauthorblockA{Benedetta Tondi}
\IEEEauthorblockA{University of Siena, Italy\\
Email:  benedettatondi@gmail.com}
}

\author{%
{Mauro Barni{\small $~^{1}$}, Kassem Kallas{\small $~^{2}$}, Ehsan Nowroozi{\small $~^{3}$}, Benedetta Tondi{\small $~^{4}$}}%
\vspace{1.6mm}\\
\fontsize{10}{10}\selectfont\itshape
Department of Information Engineering, University of Siena,
Via Roma 56, 53100 - Siena, ITALY\\
\fontsize{9}{9}\selectfont\ttfamily\upshape
%
$^{1}$\,barni@dii.unisi.it; $^{2}$\,k\_kallas@hotmail.com; \\$^{3}$\, ehsan.nowroozi65@gmail.com; $^{4}$\,benedettatondi@gmail.com
\vspace{1.2mm}\\
\fontsize{10}{10}\selectfont\rmfamily\itshape
\vspace{-0.7cm}
}

\maketitle

\begin{figure}[b]
\vspace{-0.3cm}
\parbox{\hsize}{\em
WIFS`2020, December, 6-9, 2020, New York, USA.

XXX-X-XXXX-XXXX-X/XX/\$XX.00 \ \copyright 2020 IEEE. \\ \\
*The list of authors is provided in alphabetic order. The corresponding author is Ehsan Nowroozi. \\
}\end{figure}

\begin{abstract}
Last-generation GAN models allow to generate synthetic images which are visually indistinguishable from natural ones, raising the need to develop tools to distinguish fake and natural images  thus contributing to preserve the trustworthiness of digital images.
While modern GAN models can generate very high-quality images with no visible spatial artifacts, reconstruction of consistent relationships among colour channels is expectedly more difficult. In this paper, we propose a method for distinguishing GAN-generated from natural images by exploiting inconsistencies among spectral bands, with specific focus on the generation of synthetic face images. Specifically, we use cross-band co-occurrence matrices, in addition to spatial co-occurrence matrices, as input to a CNN model, which is trained to distinguish between real and synthetic faces.
The results of our experiments confirm the goodness of our approach which outperforms a similar detection technique based on intra-band spatial co-occurrences only. The performance gain is particularly significant with regard to robustness against post-processing,
like geometric transformations, filtering and  contrast manipulations.
\end{abstract}

\IEEEpeerreviewmaketitle

\section{Introduction}
\label{sec.intro}

Thanks to the astonishing performance of deep neural networks and  Generative Adversarial Networks (GANs) \cite{GAN} in particular, even non-expert users can generate realistic fake images.
In addition to other functionalities, like image editing, modification of attributes and style transfer, GANs allow to create completely synthetic images from scratch.
The last generation of GAN architectures, in particular, can create images, noticeably face images, of very high quality, that can easily deceive a human observer \cite{brock2018large,karras2017progressive, StyleGanV1}.
%
Alongside many benign uses of this technology, the possible misuse of the synthetic contents generated by GANs represents a serious threat calling for the development of image forensic techniques capable to distinguish between real and fake (GAN-generated) images.

Several methods have been proposed in the forensic literature to reveal whether an image has been generated by a GAN or it is a natural one.
%
Most recent methods are based on CNNs and can achieve very good performance.
In particular, the approach developed in \cite{nataraj2019detecting} achieves a very good detection accuracy by computing the co-occurrence matrices from the three color image bands  and inputting them to a CNN.

The method proposed  in this paper starts from the observation that, even if
modern GAN models can generate very high quality images, with highly undetectable spatial artifacts (if any),
the reconstruction of a consistent relationship among colour bands is expected to be more difficult.
Specifically, we propose to improve the detection of GAN generated images by extending the approach proposed in \cite{nataraj2019detecting} by feeding the CNN detector with cross-band (spectral) co-occurrences, in addition to the gray-level co-occurrences computed on the single bands separately.
In our experiments, we focused on the recent StyleGAN architecture \cite{StyleGanV1,StyleGanV2} which is known to generate higher quality images with respect to previous architectures like ProGAN and StarGAN \cite{karras2017progressive,StarGAN}, making the detection task more challenging.
The proposed CNN detector can achieve almost perfect detection accuracy, moreover it exhibits a much better robustness against post-processing with respect to the intra-band method proposed in \cite{nataraj2019detecting}.
With regard to the detection of JPEG-compressed synthetic images, which is known to be a weak point of many methods proposed so far,
the model has to be re-trained by considering JPEG compressed faces.
The results we got show that the JPEG-aware model can reveal JPEG compressed GAN faces  with very high accuracy both in matched and mismatched conditions, that is, when the JPEG quality factors used for training are not equal to those used during testing. We evaluated the performance of the JPEG-aware model also in the presence of post-processing operations applied to the images before JPEG compression, to confirm that robustness against post-processing is maintained also by the JPEG-aware version of the detector.
%
%

The paper is organized as follows: in Section \ref{sec.priorArt} we briefly review the related works on GAN image detection, with specific attention to the methods based on color analysis. Then,  in Section \ref{sec.method} we present the new proposed approach. The methodology followed in our experiments is detailed in Section \ref{sec.methodology}, while the results of the experiments are reported and discussed in Section \ref{sec.exp}. We conclude the paper with some final remarks in Section \ref{sec.conc}.

\section{Related work}
\label{sec.priorArt}

Several techniques have been proposed in the literature to distinguish  GAN generated faces, and, in general, GAN images, from real ones. Some of them exploit particular facial traces (e.g., eye color, borders of face and nose \dots), that are left behind by GANs, \cite{Face2019Manip}, especially by the earlier architectures.
%
In \cite{Yang2019ExposingGF}, for instance, the authors showed that it is possible to reveal whether a face image is  GAN generated  or not by considering the locations of facial landmark points and using them to train an SVM.
Approaches that are closer to the one presented in this paper are those exploiting color information to reveal GAN contents \cite{mccloskey2018detecting,li2020detection}.
By analyzing the behavior and the color cues introduced by GAN models, two detection metrics  based on correlation between color channels and saturation are defined in \cite{mccloskey2018detecting} and considered for the detection task.
Co-occurrence-based features, often computed on residual images, have been widely used in image forensics
for detecting or localizing manipulations, e.g. in \cite{Coz2014,li2016identification,Barni2017Eusipco,Barni2018IWBF}.
More specifically,  the Subtractive Pixel Adjacency
Matrix (SPAM) features  \cite{SPAM2010}, the
rich feature model (SRMQ) \cite{Jessica2012} and the rich features models for color images (CSRMQ) \cite{goljan2014rich}, initially proposed  for  steganalysis, are usually considered by these methods.
%
The  approach in \cite{li2020detection} combines color channel analysis and SPAM-like features to perform GAN image detection.
In such a paper,
co-occurrences are extracted from truncated high-pass filtering residuals of several color components and from a truncated residual image obtained by combining the R, G and B channels. The co-occurrences are then combined into a feature vector used to train an SVM.
%

More recently, methods based on Convolutional Neural Networks (CNN) have also been proposed \cite{Marra2018GANSo,nataraj2019detecting, Xuan2019, Mansourifar2020OneShotGG, Marra2019}, achieving improved performance compared to previous methods based on standard machine learning and hand-crafted features.
An approach based on incremental learning to mitigate the need of using a large training datasets from all the different GAN architectures was proposed in \cite{Marra2019}.
%
In \cite{nataraj2019detecting} the authors have shown that improved performance can be achieved for the face GAN detection task by feeding a CNN with co-occurrence matrices computed directly on the input image, compared to the use of co-occurrences-based features extracted from noise residuals  (like in steganalytic sets of features) and SVM classification.
Inspired by this work, in this paper, we leverage on the superior performance of CNNs, 
and design a method to enforce the network to automatically learn inconsistencies among the color components by also looking at cross-band co-occurrence matrices.

\section{GAN faces detection based on Cross-band Co-Occurrences}
\label{sec.method}

It has been recently shown that GAN-generated content can be exposed by looking at inconsistencies in the pixel co-occurrences \cite{nataraj2019detecting}.
Specifically, the authors of \cite{nataraj2019detecting} propose to consider the 3 co-occurrence matrices computed on the R, G, and B channel of the images and input such tensor of 3 matrices to a CNN.
By training the network with GAN-generated images on one hand and natural images on the other, the network learns characteristic features from the co-occurrence matrices,  being able to effectively distinguish between GAN and real images, outperforming state-of-the-art methods.
%
In the following, we refer to the CNN model trained on co-occurrence matrices as described in \cite{nataraj2019detecting} as {\em CoNet}.

The method proposed in this paper starts from the observation that reconstructing consistent relationships among colour bands is expectedly more difficult for GANs.
In order to exploit the relationships among colour bands,
our proposal is to compute cross-band co-occurrences and use them as input of the CNN in addition to the spatial co-occurrences computed on the single color bands separately. We refer to the network trained in this way as {\em Cross-CoNet}.
We also expect cross-band features to be more robust to common post-processing operations, that usually focus more on spatial pixel relationships rather than on cross-band characteristics.

Formally, given an image $I(x,y,z)$ of size $N \times M \times 3$, let $R$, $G$ and $B$ denote respectively the 2D matrix for the red, green and blue channel.
Given an
offset (or displacement) $\Delta = (\Delta_x, \Delta_y)$, the spatial co-occurrence matrix of  channel $R$  is computed as 
\begin{align}
C_{\Delta}(i,j; R) & = \\ 
& \hspace{-2cm} \sum_{x=1}^N \sum_{y=1}^M \left\{\begin{array}{ll} 1 & \text{if $R(x,y) = i$ and $R(x + \Delta_x,y + \Delta_y) =
j$} \nonumber\\ 0 & \text{otherwise}\end{array}\right.
\end{align}
where $R(x,y) = I(x,y,1)$, for every $i,j \in [0,255]$.
A similar definition holds for channels G and B.

We define the {\em cross-band (spectral) co-occurrence matrix} for the pair of channels R and G as follows
\begin{align}
C_{\Delta'}(i,j; RG) & = \\
& \hspace{-2.1cm} \sum_{x=1}^N \sum_{y=1}^M \left\{\begin{array}{ll} 1 & \text{if $I(x,y,1) = i$ } \nonumber\\
& \text{and $I(x + \Delta_x',y + \Delta_y',2) = j$}
\nonumber\\ 0 & \text{otherwise}\end{array}\right.
\end{align}
for every $i,j \in [0,255]$. The offset $\Delta' = (\Delta_x', \Delta_y')$ is applied inter-channel, i.e. across the bands, instead of intra-channel.
%
%
In the cross-band case, the offset takes into account two effects: the different band and the different spatial location.
A $\Delta'$ equal to $(0,0)$ corresponds to the case of no shift in the pixel location for the cross-band co-occurrence analysis (only the spectral band changes).

A similar definition can be given for the other band combinations, namely  R and B ($C_{\Delta'}(i,j; RB)$), and G and B ($C_{\Delta'}(i,j; GB)$).

The input of {\em Cross-CoNet} network is given by the tensor $T_{\Delta,\Delta'}$  of the 6 co-occurrence matrices, of size $256\times 256 \times 6$, consisting of the 3 spatial co-occurrence matrices for channels R, G and B (separately) and the 3 cross-band co-occurrence matrices for the pairs RG, RB and  GB, namely:
\begin{align}
T_{\Delta, \Delta'}(i,j; :) = & [C_{\Delta}(i,j; R),  C_{\Delta}(i,j; G), C_{\Delta}(i,j; B), \\
& \hspace{0.25cm} C_{\Delta'}(i,j; RG), C_{\Delta'}(i,j; RB), C_{\Delta'}(i,j; GB)]. \nonumber
\end{align}
%

In  our experiments, we considered a different offset for spatial and spectral co-occurrences. Specifically, we set $\Delta = (1, 1)$  and $\Delta' = (0,0)$.

\begin{figure}[h]
	\centering
	\includegraphics[width=\columnwidth]{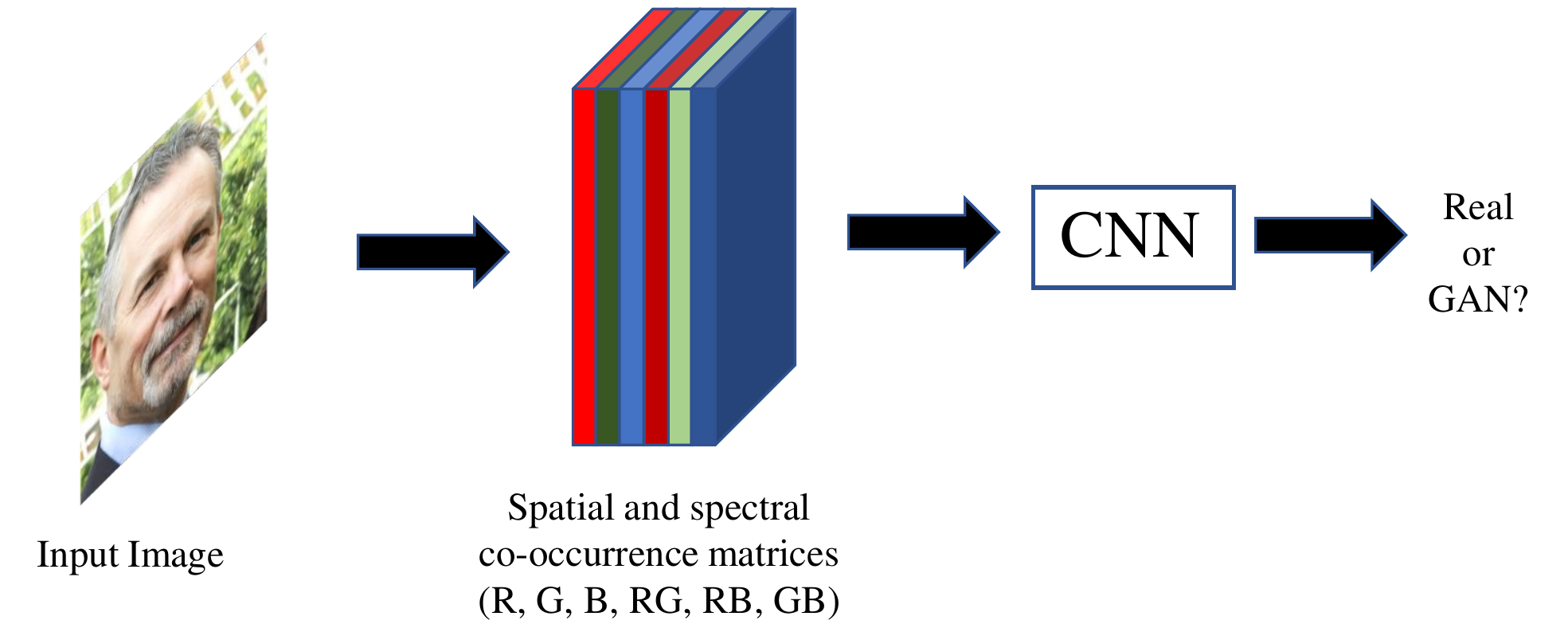}
	\caption{Scheme of the CNN-based detector considered.}
	\label{fig.sketch}
\end{figure}

The scheme of the CNN detector is illustrated in Figure \ref{fig.sketch}.

\section{Experimental Methodology}
\label{sec.methodology}

In this section, we describe the experimental methodology we followed to assess the effectiveness of our newly proposed approach.
%

\subsection{Dataset}

We consider the detection of StyleGAN2 images \cite{StyleGanV2}. StyleGAN2  has been recently proposed as an improvement of the original StyleGAN architecture \cite{StyleGanV1}, and achieves impressive results, being capably to generate synthetic images of extremely high quality.
Differently from  earlier architectures (e.g. ProGAN \cite{karras2017progressive} and StarGAN \cite{StarGAN}), face images generated with StyleGAN2 are visually indistingushable from real faces (no visual clues are left on the facial part or in the background), making their detection particularly hard.

\begin{figure}%
	\centering
	\subfloat[Real images]{{\includegraphics[width=4.3cm]{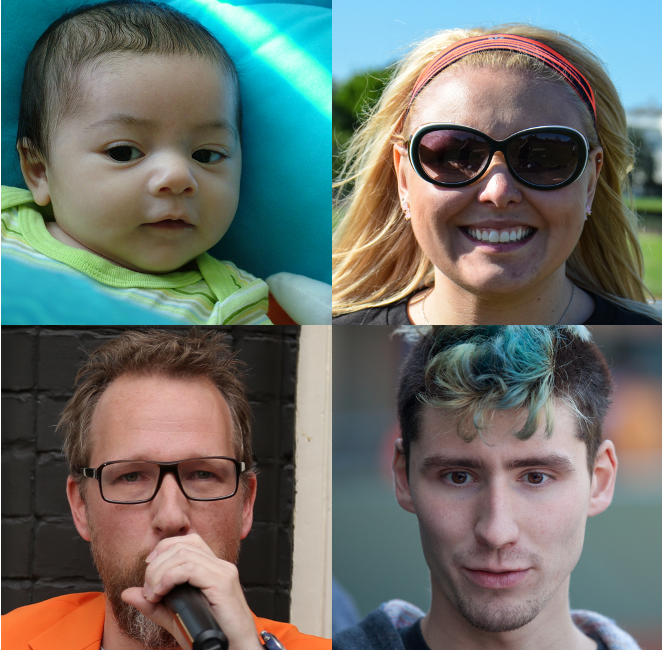} }}%
	\subfloat[Generated Images]{{\includegraphics[width=4.3cm]{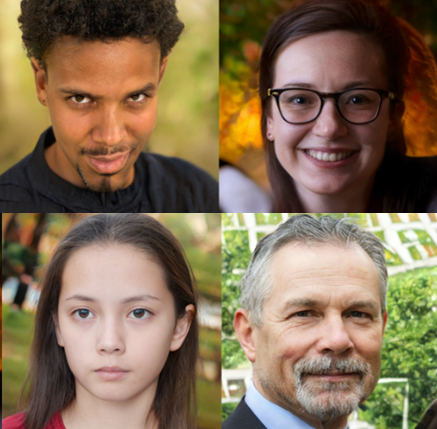} }}%
	\caption{a) Examples of real images (FFHQ database);  b) examples of images generated using StyleGan2 \cite{StyleGanV2}.}%
	\label{fig:example}%
\end{figure}

The dataset used to train StyleGAN is the Flicker-Faces-HQ (FFHQ) database \cite{karras2019style}. FFHQ consists of 70.000 high-quality human face images, produced as a benchmark for generative adversarial networks (GAN). The images are
in PNG format, with a  resolution  of $1024\times1024$, and include considerable variability in age, ethnicity, and background.
Faces covered with various accessories such as glasses, sunglasses, hats, and so on, are also included in the dataset.
%
The images were crawled from Flickr,
and automatically aligned and cropped, thus they inherit all the biases of that website.
In particular, Flickr images are JPEG compressed (even if only images with very high compression quality have been used to build the FFHQ dataset).
The FFHQ dataset is publicly available at https://github.com/NVlabs/ffhq-dataset.
Figure \ref{fig:example}(b) shows some examples of StyleGan face images.
%
%
%

\subsection{Network architecture}
\label{sec.net}

With regard to the architecture, we considered the same structure of the network already used for GAN images detection  in \cite{nataraj2019detecting}. The network consists of 6 convolutional layers followed by a single fully-connected layer.
The only modification we made regards the first input layer, since in our case we have a 6 band input instead of 3.
The structure is detailed as follows:\footnote{The stride is set to 1 for every convolution.}
\begin{itemize}
\item A convolutional layer with 32 filters of size $3 \times 3$, followed by a ReLu layer;
\item A convolutional layer with 32 filters of size $5 \times 5$ followed by a max pooling layer;
\item A convolutional layer with 64 filters of size $3 \times 3$,followed by a ReLu layer;
\item A convolutional layer with 64 filters of size $5 \times 5$ followed by a max pooling layer;
\item A convolutional layer with 128 filters of size $3 \times 3$, followed by a ReLu layer;
\item A convolutional layer with 128 filters of size $5 \times 5$ followed by a max pooling layer;
\item A dense layer (256 nodes)  followed by a sigmoid layer.
\end{itemize}

\subsection{Robustness analysis}
\label{sec.proc}

Robustness performance is evaluated against a wide variety of post-processing operations.
The robustness achieved by {\em Cross-CoNet} is compared with that achieved by the network in \cite{nataraj2019detecting} ({\em CoNet}).
Specifically, we consider geometric transformations (resizing, rotation, zooming, and cropping), filtering operations  (median filtering, and blurring) and adjustments of the contrast (gamma correction and adaptive histogram equalization, referred as to AHE).
For resizing (downscaling),  0.9, 0.8, and 0.5 are considered as a scaling factors, while the zooming (upscaling) is applied with factors  1.1, 1.2, and 1.9. The bicubic interpolation is considered for all the rescaling operation. For rotation, we considered angles of 5, 10, 45 degrees with bicubic interpolation.
%
%
Cropping is applied by considering a $880 \times 880$ area.
With regard to filtering, we set the window size for both median filtering and blurring to $3 \times 3$ and $5 \times 5$. Finally, for noise addition,  we consider Gaussian noise
with standard deviations  $\sigma = 0.5$, $ 0.8$
and $2$ with zero mean.
For gamma correction, we let $\gamma \in  \{ 0.8, 0.9, 1.2\}$, while for adaptive Histogram Equalization (in particular, its refined version, Contrast Limited, impelementation CLAHE \cite{ZUIDERVELD1994474}), the clip-limit parameter is set to 1.0 . Finally, robustness is assessed against blurring with window size $3 \times 3$ followed by sharpening with kernel [[-1, -1, -1], [-1, 9, -1], [-1, -1, -1]].

The performance of the JPEG-aware version of the {\em Cross-CoNet} detector is also assessed in the presence of post-processing operations applied prior to compression.
Specifically, a JPEG compressed version of the processed images for the case of resizing (with scale factor 0.9), median filtering ($3\times 3$) and noise addition ($\sigma = 2$) was produced, the compression being applied with different QFs.


\section{Experimental Analysis}
\label{sec.exp}

\subsection{Setting}
\label{sec.setting}


For our experiments, we considered a total of 20000 real (from FFHQ) and 20000 GAN-generated images, split as follows for both real and GANs:
12000 (60\%) were considered for training, 4000 (20\%) for validation, and 4000 (20\%) for testing.
As optimiser, we employed the stochastic gradient descent (SGD) \cite{SGD}, with learning rate 0.01 and momentum 0.9 \cite{SGD}. The batch size was set to 40.
The network was trained for 40 epochs.
Network training and testing have been implemented in TensorFlow via the Keras API.
For a fair comparison, we used the same setting to train the {\em CoNet} model.

For the robustness experiments, the images were post-processed with the OpenCV library  in Python. For each processing type and parameters, the tests were run on 2000 images per class from the test set.



To get the JPEG-aware {\em Cross-CoNet} model for the detection of GAN-generated face images, the following quality factors were considered for JPEG-aware training: QF $\in \{75, 80, 85, 90, 95\}$.
Specifically, for both the real and GAN classes, 5000 images for each quality factor have been considered for training for a total of 25000 images. 5000 images were used for validation and for testing (1000 per each QF). The model was trained from scratch for 40 epochs by using the SGD optimizer.
The same number of test images were considered to test the robustness of the JPEG-aware model against post-processing.


\subsection{Results}

\subsubsection{Performance of Cross-CoNet}

The test accuracy achieved by {\em Cross-CoNet} on the test set for the StyleGAN2 detection task is  99.70\%, which is slightly better than the accuracy reached by {\em CoNet}, which is equal 98.15\% (see Table \ref{tab0}, first two columns). 
This represents a minor improvement since {\em CoNet} already works very well for the plain GAN detection task.

The main advantage of {\em Cross-CoNet} over {\em CoNet} is in the increased robustness against post-processing.
%
%
Table \ref{tab1} reports the accuracy of the tests carried out under various post-processing operations.
In all the cases the robustness achieved by {\em Cross-CoNet} is much higher, even when the post-processing applied to the images is strong.
The worst scenario corresponds to the case of AHE and blurring followed by sharpening, when the accuracy of {\em Cross-CoNet} drops to 75\% or slightly less. 
By looking at the results of {\em CoNet}, we observe that in many cases the accuracy of this network is close or even equal to 50\%. The reason is that in these cases, after post-processing, almost all the GAN images  are classified are real, meaning that the artifacts that the network relies on to classify the image as GAN are washed out by post-processing. Even in the case of {\em Cross-CoNet}, the few errors are made mainly on images for the GAN class, while real images are always classified correctly after processing.
Despite a small loss, the performance of {\em Cross-CoNet}  are very good confirming that
by looking at the cross-band co-occurrences, the network can learn stronger features characterizing GAN faces thus resulting in a model that is resistant to subsequent processing.
Moreover, most of the processing operations alter the spatial relationships among pixels while they do not alter much the intra-channel relationships, thus providing another reason for the improved robustness of {\em Cross-CoNet} against post-processing.

\begin{table}[h!]
	\renewcommand\arraystretch{1.2}
	\centering
	\caption{
Accuracy of  {\em Cross-CoNet} and  {\em CoNet}.}
	\label{tab0}
	\begin{tabular}{|c|c|c|c|}
\hline
\multicolumn{2}{|c|}{\textbf{Unaware model}} & \multicolumn{2}{|c|}{\textbf{JPEG-aware model}}\\ \hline
		\textbf{Cross-CoNet} & \textbf{CoNet} &  \textbf{Cross-CoNet} & \textbf{CoNet}\\ \hline
		99.70\% &  98.15\% & 95.50\% & 94.83\% \\ \hline
	\end{tabular}
		
\end{table}

\begin{table}[h!]
	\renewcommand\arraystretch{1.2}
	\centering
	\caption{
Robustness performance in the presence of post-processing.  The accuracy values are reported. }
	\label{tab1}
	\begin{tabular}{|l|c|c|c|}
		
		\hline
		
		\textbf{Processing} & \textbf{parameter}  & \textbf{Cross-CoNet} & \textbf{CoNet \cite{nataraj2019detecting}}\\ \hline
		
		\multirow{2}{*}{Median filtering} & $3 \times 3$ &  96.25\% & 50.00\% \\ \cline{2-4}
		
		 & $5\times 5$ &  90.35\%  & 50.00\% \\ \hline
		
		
		 \multirow{3}{*}{Noise} & 0.5  & 99.95\% & 86.40\% \\ \cline{2-4}
		
		 & 0.8  & 99.55\% &  64.10\% \\ \cline{2-4}
		
		 & 2  &90.70\%  &  50.00\% \\ \hline
		
		AHE &  -  & 75.00\% &  50.00\% \\ \hline
		
		
		\multirow{3}{*}{Gamma correction} & 0.9 &99.65\% &  55.90\% \\ \cline{2-4}
		
		 & 0.8  & 82.50\% &  50.80\% \\ \cline{2-4}
		
		
		& 1.2  & 91.70\%  &  50.15\%\\ \hline
		
		\multirow{2}{*}{Average blurring} & $3\times 3$  & 92.85\% &  72.50\% \\ \cline{2-4}
		
		& $5\times 5$ & 85.30\%  &  54.10\% \\ \hline
		
		\multirow{3}{*}{Resize} &  0.9  & 99.73\% &  90.78\%  \\ \cline{2-4}
		
		&0.8  & 99.50\% &  76.65\% \\ \cline{2-4}
		
		&0.5   & 81.50\% &  50.05\% \\ \hline
		
		\multirow{3}{*}{Zooming}  &  1.1  & 99.60\% &  94.95\% \\ \cline{2-4}
		
		 & 1.2   & 99.45\% &  89.95\% \\  \cline{2-4}
		
		  &1.9   & 98.60\% &  57.65\% \\ \hline
		
		\multirow{3}{*}{Rotation} & 5  & 99.45\% &  93.10\% \\ \cline{2-4}
		
		 & 10  & 99.50\% &  93.65\% \\ \cline{2-4}
		
		 & 45  & 99.50\% &  71.90\% \\ \hline
		
%
		
		Crop & -   & 99.80\% & 92.60\%\\ \hline
		
		\makecell{Blurring followed by \\sharpening} & -   & 73.60\% & 50.00\%\\ \hline

	\end{tabular}

\end{table}

As a further result, we checked that the performance of both {\em CrossNet} and {\em CoNet}  decrease in the presence of JPEG compression. Specifically, the detection accuracy is already lower than 80\% when $QF =$ 95, and falls below 70\% when  $QF =$ 85.
However, such a loss of performance under JPEG compression is not surprising since most of the (if not all) real images in the FFHQ dataset have been JPEG compressed at least once and then they exhibit traces of compression (although these traces can be weak, since the quality of  these images is high). The GAN images, instead, do not exhibit those traces. Then, compression artefacts can be indirectly associated by the network to the class of real images.

\subsubsection{Performance of JPEG-aware Cross-CoNet}

Based on our experiments, the JPEG-aware  {\em Cross-CoNet} yields an average accuracy of 95.5\% on the test set of JPEG real and GAN faces,  under the same values of the quality factors $\{75, 80, 85, 90, 95\}$ considered for training. By training  a JPEG-aware version of {\em CoNet} network we obtained a  test accuracy of 94.83\%, with a similar small gain of {\em Cross-CoNet} over {\em CoNet} than in the unaware case  (see rightmost columns in Table \ref{tab0}).
%
%

We  verified that the model trained on the selected values of quality factors
generalizes well to other compression qualities. Results of the tests carried out under both matched and mismatched values of the QF are reported in Table \ref{tabJPEG}. The average loss of accuracy when mismatched quality factors are considered (in the range $\{73, 77, 83, 87, 93,  97\}$) is less than 1\%, thus showing that the generalization capabilities of the network are very good.

With regard to the performance of the {\em JPEG-aware Cross-CoNet} model in the presence of processing operators applied before the final compression, we verified that the reduction of the accuracy achieved by the model is very small.  Results are reported in Table \ref{tabJPEG-PP} for the case of median filtering (window size $3\times 3$), resizing (scale factor 0.9), and noise addition ($\sigma = 2$).  Specifically, the average accuracy in the range $\{73, 75, 77, 80, 83, 85, 87, 90, 93, 95, 97\}$ is 94.20\%, 93.10\% and 93\%  in the case of median filtering, resizing, and noise addition respectively.
These results show that the good robustness of {\em Cross-CoNet} is maintained by the JPEG-aware version of the model, thus confirming that relying on the inconsistencies among the color bands by means of the cross-band co-occurrences is an effective way to distinguish between real and GAN-generated images.

\begin{table}[h!]
	\renewcommand\arraystretch{1.2}
	\centering
	\setlength{\tabcolsep}{3pt}
	\caption{Accuracy of the {\em JPEG-aware Cross-CoNet} detector under matched and mismatched values of $QF$. Matched values are highlighted in bold.}
	\label{tabJPEG}
%
%
%
%
%
%
%
%
%
%
%
%
	\begin{tabular}{|c|c|c|c|c|c|c|}
		
		\hline
		
		\textbf{QF} & 73  & \textbf{75} & 77   & \textbf{80}  & 83 & \textbf{85}  \\ \hline
		
		Accuracy  &  94.30\% & 94.70\% & 94.90\%  &  95.40\% & 94.80\% & 95.20\% \\ \hline
		
		\textbf{QF} & 87  & \textbf{90} & 93 &  \textbf{95}  & 97 & ----  \\ \hline
		
		Accuracy  &  95.40\% & 94.80\% & 95.50\% &  95.00\% & 94.70\% & ---- \\ \hline

	\end{tabular}
	
\end{table}

\begin{table}[h!]
\renewcommand\arraystretch{1.4}
	\centering
	\caption{Robustness performance (Accuracy) of {\em JPEG-aware Cross-CoNet} in the presence of post-processing. }
	\label{tabJPEG-PP}
	\begin{tabular}{|c|c|c|c|}	
		\hline
\textbf{QF}  & \textbf{median filtering} & \textbf{resizing} & \textbf{noise addition} \\ \hline
73 & 93.90\% & 92.20\% & 93.10\% \\ \hline
75 & 94.20\% & 93.20\% & 92.90\%  \\ \hline
77 & 94.30\% & 93.80\%& 93.40\% \\ \hline
80 & 94.10\%&  94.50\%& 93.60\%		\\ \hline
83 & 94.10\% &  94.10\%& 93.00\%		\\ \hline
85 & 93.80\% & 94.20\%& 93.50\% \\ \hline
87 & 94.20\% & 94.50\%& 93.50\% \\ \hline
90 & 94.20\% & 94.50\% & 93.20\%  \\ \hline
93 & 94.20\%& 94.40\% & 91.20\%  \\ \hline
95 & 94.20\% & 94.90\%& 88.40\%  \\ \hline
97 & 94.50\% & 94.70\%& 88.00\%  \\ \hline
	\end{tabular}

\end{table}

\section{Conclusions}
\label{sec.conc}

We have introduced a CNN method for distinguishing GAN-generated images from real images, with particular focus on synthetic face images detection. The proposed method exploits inconsistencies among spectral bands.  Cross-band co-occurrence matrices are used, in addition to pixel co-occurrence matrices, to train a CNN model to extract discriminative features for the  real and GAN classes. The experimental analysis we carried out shows the good performance of the proposed method and the improved robustness that can be obtained against various post-processing compared to the case where only spatial co-occurrences are considered to train the detector, confirming the benefit of looking at the relationships among color bands.

Future research will be devoted to investigate the performance of {\em Cross-CoNet}
under intentional attacks, purposely modifying the band relationships in order to confuse the detector, both in a white-box and black-box setting.
In this regard, it would be interesting to assess the performance of the network against
an informed attacker who runs adversarial examples against the proposed CNN, by approximating the co-occurrence computation via convolutions  \cite{cozzolino2017recasting} and then backpropagating the gradient to the pixel domain.
Another direction for future research is to investigate the generalization capability of \textit{Cross-CoNet} to unknown datasets. 
Finally, recent work has shown CNN-based image manipulation detectors are vulnerable to \textit{rebroadcast} attack \cite{Farid2018Rebroadcast}. Hence, it would also be interesting to study the performance of \textit{Cross-CoNet} and \textit{CoNet} when facing with a print-and-scan attack.

\section*{Acknowledgements}

This work has been partially supported by  the PREMIER project, funded
by the Italian Ministry of University and Research (MUR) under
the PRIN 2017 2017Z595XS-001 program. 


\bibliographystyle{IEEEtran}

\bibliography{Ref}

\begin{thebibliography}{10}
\providecommand{\url}[1]{#1}
\csname url@samestyle\endcsname
\providecommand{\newblock}{\relax}
\providecommand{\bibinfo}[2]{#2}
\providecommand{\BIBentrySTDinterwordspacing}{\spaceskip=0pt\relax}
\providecommand{\BIBentryALTinterwordstretchfactor}{4}
\providecommand{\BIBentryALTinterwordspacing}{\spaceskip=\fontdimen2\font plus
\BIBentryALTinterwordstretchfactor\fontdimen3\font minus
  \fontdimen4\font\relax}
\providecommand{\BIBforeignlanguage}[2]{{%
\expandafter\ifx\csname l@#1\endcsname\relax
\typeout{** WARNING: IEEEtran.bst: No hyphenation pattern has been}%
\typeout{** loaded for the language `#1'. Using the pattern for}%
\typeout{** the default language instead.}%
\else
\language=\csname l@#1\endcsname
\fi
#2}}
\providecommand{\BIBdecl}{\relax}
\BIBdecl

\bibitem{GAN}
I.~Goodfellow, J.~Pouget-Abadie, M.~Mirza, B.~Xu, D.~Warde-Farley, S.~Ozair,
  A.~Courville, and Y.~Bengio, ``Generative adversarial nets,'' in
  \emph{Advances in Neural Information Processing Systems 27}, Z.~Ghahramani,
  M.~Welling, C.~Cortes, N.~D. Lawrence, and K.~Q. Weinberger, Eds.\hskip 1em
  plus 0.5em minus 0.4em\relax Curran Associates, Inc., 2014, pp. 2672--2680.

\bibitem{brock2018large}
A.~Brock, J.~Donahue, and K.~Simonyan, ``Large scale gan training for high
  fidelity natural image synthesis,'' \emph{arXiv preprint arXiv:1809.11096},
  2018.

\bibitem{karras2017progressive}
T.~Karras, T.~Aila, S.~Laine, and J.~Lehtinen, ``Progressive growing of gans
  for improved quality, stability, and variation,'' \emph{arXiv preprint
  arXiv:1710.10196}, 2017.

\bibitem{StyleGanV1}
T.~{Karras}, S.~{Laine}, and T.~{Aila}, ``A style-based generator architecture
  for generative adversarial networks,'' in \emph{2019 IEEE/CVF Conference on
  Computer Vision and Pattern Recognition (CVPR)}, June 2019, pp. 4396--4405.

\bibitem{nataraj2019detecting}
L.~Nataraj, T.~M. Mohammed, B.~Manjunath, S.~Chandrasekaran, A.~Flenner, J.~H.
  Bappy, and A.~K. Roy-Chowdhury, ``Detecting gan generated fake images using
  co-occurrence matrices,'' \emph{Electronic Imaging}, vol. 2019, no.~5, pp.
  532--1, 2019.

\bibitem{StyleGanV2}
T.~Karras, S.~Laine, M.~Aittala, J.~Hellsten, J.~Lehtinen, and T.~Aila,
  ``Analyzing and improving the image quality of stylegan,'' \emph{ArXiv}, vol.
  abs/1912.04958, 2019.

\bibitem{StarGAN}
\BIBentryALTinterwordspacing
Y.~Choi, M.~Choi, M.~Kim, J.~Ha, S.~Kim, and J.~Choo, ``Stargan: Unified
  generative adversarial networks for multi-domain image-to-image
  translation,'' \emph{CoRR}, vol. abs/1711.09020, 2017. [Online]. Available:
  \url{http://arxiv.org/abs/1711.09020}
\BIBentrySTDinterwordspacing

\bibitem{Face2019Manip}
F.~{Matern}, C.~{Riess}, and M.~{Stamminger}, ``Exploiting visual artifacts to
  expose deepfakes and face manipulations,'' in \emph{IEEE Winter Applications
  of Computer Vision Workshops (WACVW)}, 2019, pp. 83--92.

\bibitem{Yang2019ExposingGF}
X.~Yang, Y.~Li, H.~Qi, and S.~Lyu, ``Exposing gan-synthesized faces using
  landmark locations,'' \emph{Proceedings of the ACM Workshop on Information
  Hiding and Multimedia Security}, 2019.

\bibitem{mccloskey2018detecting}
S.~McCloskey and M.~Albright, ``Detecting gan-generated imagery using color
  cues,'' \emph{arXiv preprint arXiv:1812.08247}, 2018.

\bibitem{li2020detection}
H.~Li, B.~Li, S.~Tan, and J.~Huang, ``Identification of deep network generated
  images using disparities in color components,'' \emph{Signal Processing}, p.
  107616, 2020.

\bibitem{Coz2014}
D.~{Cozzolino}, D.~{Gragnaniello}, and L.~{Verdoliva}, ``Image forgery
  detection through residual-based local descriptors and block-matching,'' in
  \emph{2014 IEEE International Conference on Image Processing (ICIP)}, Oct
  2014, pp. 5297--5301.

\bibitem{li2016identification}
H.~Li, W.~Luo, X.~Qiu, and J.~Huang, ``Identification of various image
  operations using residual-based features,'' \emph{IEEE Transactions on
  Circuits and Systems for Video Technology}, vol.~28, no.~1, pp. 31--45, 2016.

\bibitem{Barni2017Eusipco}
M.~{Barni}, E.~{Nowroozi}, and B.~{Tondi}, ``Higher-order, adversary-aware,
  double jpeg-detection via selected training on attacked samples,'' in
  \emph{25th European Signal Processing Conference (EUSIPCO)}, 2017, pp.
  281--285.

\bibitem{Barni2018IWBF}
------, ``Detection of adaptive histogram equalization robust against jpeg
  compression,'' in \emph{International Workshop on Biometrics and Forensics
  (IWBF)}, 2018, pp. 1--8.

\bibitem{SPAM2010}
T.~{Pevny}, P.~{Bas}, and J.~{Fridrich}, ``Steganalysis by subtractive pixel
  adjacency matrix,'' \emph{IEEE Transactions on Information Forensics and
  Security}, vol.~5, no.~2, pp. 215--224, June 2010.

\bibitem{Jessica2012}
J.~{Fridrich} and J.~{Kodovsky}, ``Rich models for steganalysis of digital
  images,'' \emph{IEEE Transactions on Information Forensics and Security},
  vol.~7, no.~3, pp. 868--882, June 2012.

\bibitem{goljan2014rich}
M.~Goljan, J.~Fridrich, and R.~Cogranne, ``Rich model for steganalysis of color
  images,'' in \emph{IEEE International Workshop on Information Forensics and
  Security (WIFS)}, 2014, pp. 185--190.

\bibitem{Marra2018GANSo}
F.~{Marra}, D.~{Gragnaniello}, D.~{Cozzolino}, and L.~{Verdoliva}, ``Detection
  of gan-generated fake images over social networks,'' in \emph{IEEE Conference
  on Multimedia Information Processing and Retrieval (MIPR)}, 2018, pp.
  384--389.

\bibitem{Xuan2019}
X.~Xuan, B.~Peng, W.~Wang, and J.~Dong, ``On the generalization of gan image
  forensics,'' in \emph{Biometric Recognition}, Z.~Sun, R.~He, J.~Feng,
  S.~Shan, and Z.~Guo, Eds.\hskip 1em plus 0.5em minus 0.4em\relax Cham:
  Springer International Publishing, 2019, pp. 134--141.

\bibitem{Mansourifar2020OneShotGG}
H.~Mansourifar and W.~dong Shi, ``One-shot gan generated fake face detection,''
  \emph{ArXiv}, vol. abs/2003.12244, 2020.

\bibitem{Marra2019}
F.~{Marra}, C.~{Saltori}, G.~{Boato}, and L.~{Verdoliva}, ``Incremental
  learning for the detection and classification of gan-generated images,'' in
  \emph{2019 IEEE International Workshop on Information Forensics and Security
  (WIFS)}, 2019, pp. 1--6.

\bibitem{karras2019style}
T.~Karras, S.~Laine, and T.~Aila, ``A style-based generator architecture for
  generative adversarial networks,'' in \emph{Proceedings of the IEEE
  conference on computer vision and pattern recognition}, 2019, pp. 4401--4410.

\bibitem{ZUIDERVELD1994474}
K.~Zuiderveld, ``Viii.5. - contrast limited adaptive histogram equalization,''
  in \emph{Graphics Gems}, P.~S. Heckbert, Ed.\hskip 1em plus 0.5em minus
  0.4em\relax Academic Press, 1994, pp. 474 -- 485.

\bibitem{SGD}
\BIBentryALTinterwordspacing
S.~Ruder, ``An overview of gradient descent optimization algorithms,''
  \emph{CoRR}, vol. abs/1609.04747, 2016. [Online]. Available:
  \url{http://arxiv.org/abs/1609.04747}
\BIBentrySTDinterwordspacing

\bibitem{cozzolino2017recasting}
D.~Cozzolino, G.~Poggi, and L.~Verdoliva, ``Recasting residual-based local
  descriptors as convolutional neural networks: an application to image forgery
  detection,'' in \emph{Proceedings of the 5th ACM Workshop on Information
  Hiding and Multimedia Security}, 2017, pp. 159--164.

\bibitem{Farid2018Rebroadcast}
S.~{Agarwal}, W.~{Fan}, and H.~{Farid}, ``A diverse large-scale dataset for
  evaluating rebroadcast attacks,'' in \emph{2018 IEEE International Conference
  on Acoustics, Speech and Signal Processing (ICASSP)}, 2018, pp. 1997--2001.

\end{thebibliography}

\end{document}